\documentclass{article}
\usepackage{spconf,amsmath,graphicx}
\usepackage{multirow}
\usepackage{makecell}
\usepackage[colorlinks,
            linkcolor=black,
            anchorcolor=black,
            citecolor=green]{hyperref}
\usepackage{color}

\title{Pixel-level self-paced learning for super-resolution}
\name{Wei Lin, Junyu Gao, Qi Wang$^{*}$, Xuelong Li \thanks{$^{*}$Qi Wang is the corresponding author. This work was supported by the National Key R\&D Program of China under Grant 2018YFB1107403, National Natural Science Foundation of China under Grant U1864204, 61773316, U1801262, 61871470 and 61761130079.}}
\address{School of Computer Science and Center for OPTical IMagery Analysis and Learning (OPTIMAL), \\
Northwestern Polytechnical University, Xi'an 710072, Shaanxi, P. R. China}
\begin{document}
%
\maketitle
\begin{abstract}
Recently, lots of deep networks are proposed to improve the quality of predicted super-resolution (SR) images, due to its widespread use in several image-based fields. However, with these networks being constructed deeper and deeper, they also cost much longer time for training, which may guide the learners to local optimization. To tackle this problem, this paper designs a training strategy named Pixel-level Self-Paced Learning (PSPL) to accelerate the convergence velocity of SISR models. PSPL imitating self-paced learning gives each pixel in the predicted SR image and its corresponding pixel in ground truth an attention weight, to guide the model to a better region in parameter space. Extensive experiments proved that PSPL could speed up the training of SISR models, and prompt several existing models to obtain new better results. Furthermore, the source code is available at \url{https://github.com/Elin24/PSPL}


\end{abstract}
\section{Introduction}
\label{sec:intro}

The main aim of single image super-resolution (SISR) is to reconstruct a new high-resolution (HR) image with excellent quality from a low-resolution (LR) image. It is widely applied in processing medical images~\cite{wang2019transform}, satellite images~\cite{huang2019hyperspectral}, renovating old pictures and so on~\cite{yang2019deep, wang2014learning, wang2018detecting}. As a classical task in computer vision, SISR is a challenging problem since it is a one-to-many mapping, which means an LR image could correspond to multiple HR image~\cite{yang2019deep}.

To obtain HR image with more delicate details, plenty of algorithms~\cite{DBLP:conf/icassp/ChangC19, DBLP:conf/icassp/BareLYFY18, dong2014learning, LapSRN, kim2016accurate, tai2017memnet, lim2017enhanced, haris2018deep, tai2017image}
are proposed and achieve promising results. Especially in the last half decade, the development of deep neural networks leads to a tremendous leap in this field. The pioneering deep-learning-based work is SRCNN~\cite{dong2014learning}, which only contains three convolutional layers, but establishes advanced HR image compared with traditional interpolation-based and example-based methods~\cite{tang2016example}. Different from it, VDSR~\cite{kim2016accurate} adopts a very deep structure with 20 layers, and experiments illustrate it can achieve better performance. However, the repetitive architecture in VDSR is too plain to construct a far deeper network. A representative solution for this problem is residual learning. A typical algorithm adopting it is SRResNet~\cite{ledig2017photo}, which takes 16 residual units as backbone. Based on the success of SRResNet, Lim~\textit{et al.} proposes EDSR~\cite{lim2017enhanced} to further increase the depth of networks and achieve state-of-the-art results. EDSR firstly removes all batch normalization (BN) layer in its residual units and then increases both its depth and width. For the former, it contains 32 residual units that are twice of SRResNet; for the latter, each residual unit in EDSR outputs 256 channels, which is only 64 in SRResNet.

\begin{figure}
	\centering
	\includegraphics[width=0.48\textwidth]{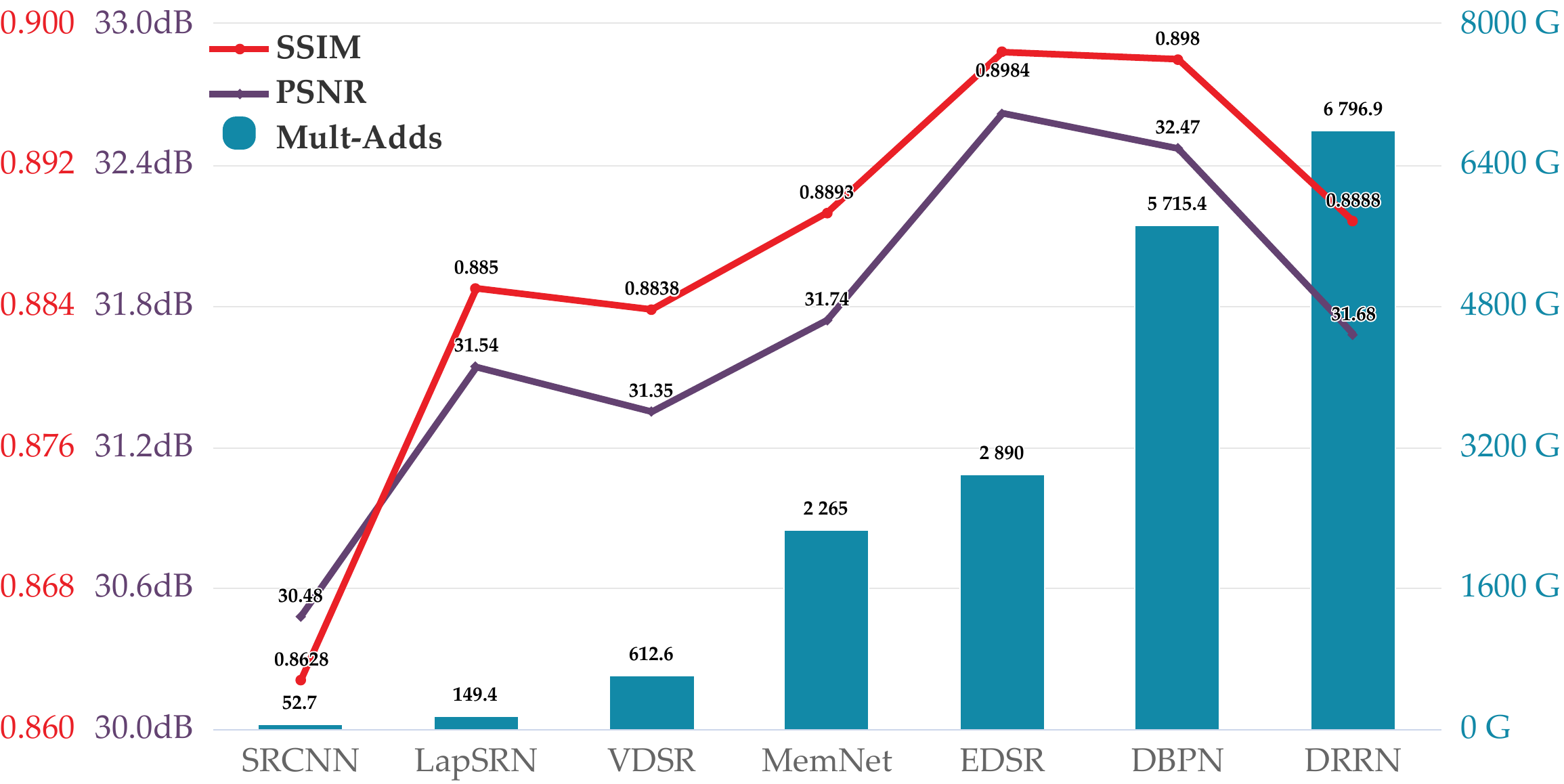}
    \caption{\textbf{Comparison of some popular deep SISR models.} SSIM and PSRN are employed for quality assessment; Mult-Adds is the number of multiplication and addition for predicting an SR image with fixed size (720p here). }
	\label{fig:cppm}
\end{figure}

These deep models indeed get outstanding performance, but they are too large and intricate to be trained efficiently. To specifically explain this point, Fig.~\ref{fig:cppm} demonstrates a survey of some popular deep networks. As shown in it, even the record is refreshed continuously, the computation complexity also keeps growing. Half of them calculate more than 2,000 times to output a regular SR image, which indicates that they may consume much more time on training. A typical example is that Lim~\textit{et al.} costs 7 days to train EDSR~\cite{lim2017enhanced}. Moreover, even though a model takes longer training time, it still does not produce better results (DBPN and DRRN). So the problem is, how to design a training strategy that could accelerate the training speed of these models, and further enhance their performance?

\begin{figure}
	\centering
	\includegraphics[width=0.48\textwidth]{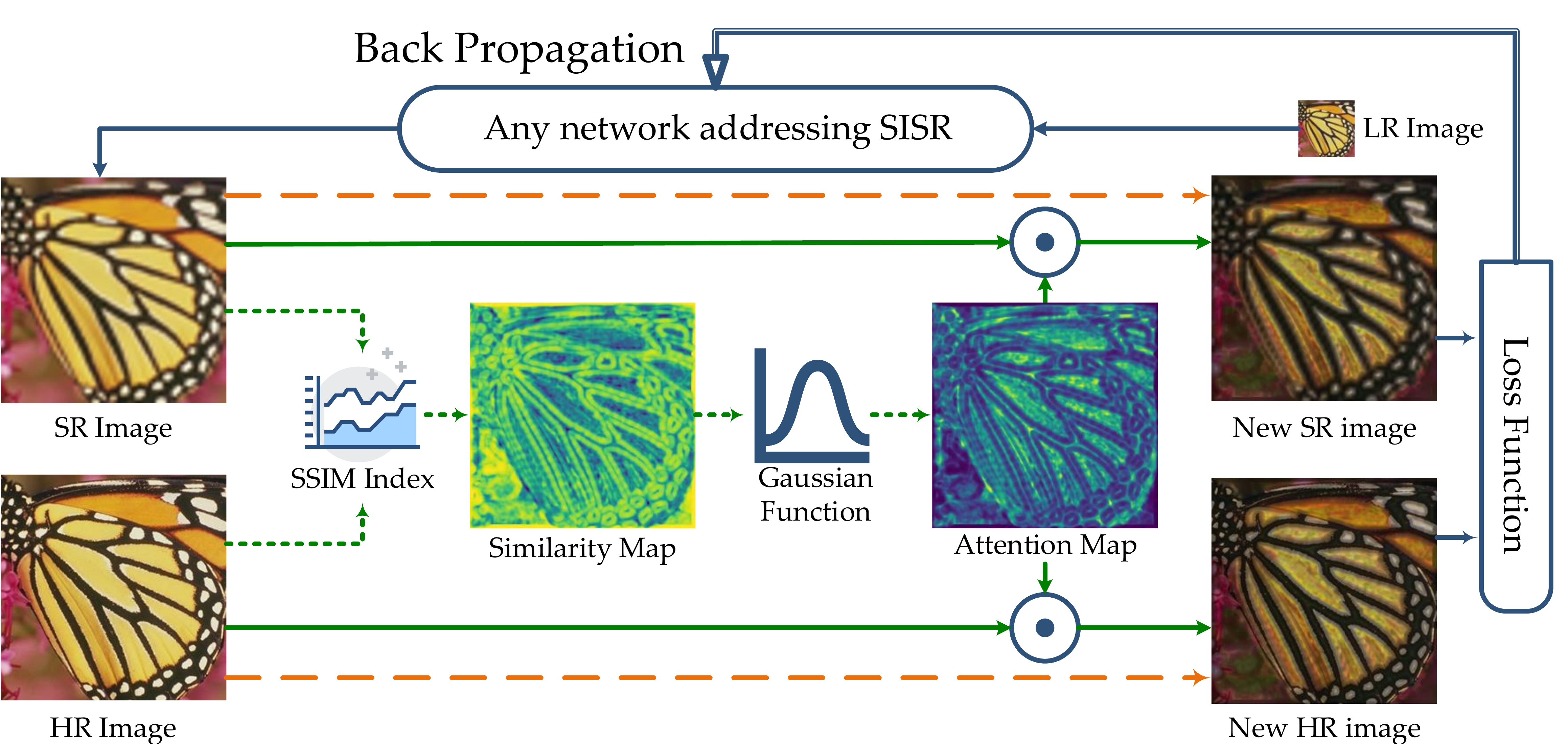}
    \caption{\textbf{PSPL algorithm flow diagram.} These arrows imply how data flows. Blue arrows represent these data flows outside PSPL. Orange dash arrows represent how loss is calculated in common training process. Green arrows display how data flow in PSPL.
    }
	\label{fig:fosflow}
\end{figure}

To remedy the above problem, this paper proposes a training strategy named Pixel-level Self-Paced Learning (PSPL). It is inspired by traditional sample-level self-paced learning (SPL), in which each sample is given a learning weight, and leads the learner to learn more efficient characteristics for quicker convergence and better generalization. However, sample-level SPL is not suitable for SISR, since SISR focus more on local pixel value. This is also the reason why this paper explores the feasibility of pixel-level SPL. Fig.~\ref{fig:fosflow} illustrates how data flow is cut off and rearranged by PSPL. To make it easier to follow, each training epoch is divided into following four steps to be introduced. Given a pair of LR image and HR image, firstly an SR image is predicted by a model when inputting the LR one. After that, a similarity map is produced according to the SR and HR image. Thirdly, based on the similarity map, an attention map is generated, and the attention map give more attention to these pairs of pixels with larger difference. At last, new SR(HR) image is obtained by doing element-wise multiplication between the attention map and original SR(HR) image, and the optimization of the SISR model is impacted by replacing original SR(HR) image with new SR(HR) image when calculating loss. Moreover, in the entire training process, all values in the attention map are going to be close to a constant as the training step increased, which means that the utility of PSPL is gradually diminished in the entire training process. 

The name of PSPL comes from two aspects. One is that all attention weights gradually decay with training time goes on, which is similar to the process that self-paced learning increases the difficult of learned objects; the other one is that PSPL allocates attention weight to each pixel in images, which is different from assigning weight for sample in traditional sample-level self-paced learning.




\section{Proposed Method}
\label{sec:fos}

As described in Sec.~\ref{sec:intro}, the whole strategy are devided into four parts, which are given shorthand terms as (a) super-resolution image generation, (b) similarity map production, (c) attention map calculation, and (d) loss function.

\subsection{Super-resolution Image Generation}
\label{subsec:sig}

In SISR, given a super resolution scale $s$, a typical model $\mathcal{F}_s$ works like below:
\begin{equation}
    I_{sr} = \mathcal{F}_s(I_{lr}) \text{,} \label{eq:sisr}
\end{equation}
in which $I_{lr}$ is the input LR image, $I_{sr}$ is the predicted SR image. During training, the HR image corresponding to $I_{lr}$ is denoted as $I_{hr}$, and the parameters in $\mathcal{F}_s$ is optimized by:
\begin{equation}
    \min_{\Theta} \mathcal{L}(I_{sr}, I_{hr}) \textit{,} \label{eq:opt}
\end{equation}
in which $\mathcal{L}$ is loss function that will be described in Sec.~\ref{subsec:lf}. $I_{sr}$ and $I_{hr}$ have the same size, and their width and height is $s$ times longer than $I_{lr}$.

\subsection{Similarity Map Production}
\label{subsec:psm}

After $I_{sr}$ is figured out, a similarity map $M_s$ is produced. It is used to measure the similarity of each corresponding local region in $I_{sr}$ and $I_{hr}$. The production can be formulated by:
\begin{equation}
    M_s = \mathcal{S}(I_{sr}, I_{hr}) \textit{,} \label{eq:si}
\end{equation}
and $M_s$ has the same size as $I_{sr}$ or $I_{hr}$. PSPL adopts structural similarity (SSIM) index~\cite{wang2004image} as $\mathcal{S}$. 
To be specific, $M_s$ is obtained through performing SSIM index on a series of patches, and these patches are obtained by using a local window with fixed size to scan the entire image pixel-by-pixel. 

Given the $i^{th}$ pair of patches $(p_{s}^i, p_{h}^i)$, which belong to $I_{sr}$ and $I_{hr}$ respectively, SSIM consists of following two steps to deal with them and then produce their similarity $m_s^i$. For convenience, the following part omits the superscript $i$.

\textbf{(1)} Apply a weighting function to $p_s$ and $p_h$ respectively to emphasize their center pixel and avoid blocking artifacts:
\begin{equation}
    \hat{p} = G \odot p \textit{,} \label{eq:gpq}
\end{equation}
in which $p \in \{ p_s, p_h\}$, $G$ is a circular-symmetric Gaussian weighting matrix, with the same size as $p$ and normalized to unit sum; $\odot$ means element-wise multiplication. After this step, new patches $(\hat{p_s}, \hat{p_h})$ are obtained.

\textbf{(2)} Calculate each individual SSIM value through:
\begin{equation}
    m_s = \frac {(2\mu_{\hat{p_s}}\mu_{\hat{p_h}} + C_1)(2\sigma_{\hat{p_s}\hat{p_h}} + C_2)} {(\mu_{\hat{p_s}}^2 + \mu_{\hat{p_h}}^2 + C_1)(\sigma_{\hat{p_s}}^2 + \sigma_{\hat{p_h}}^2 + C_2)} 
    \text{,} \label{eq:ssim}
\end{equation}
where $\mu_{\hat{p_s}}$ and $\sigma_{\hat{p_s}}^2$ are the average and variance of $\hat{p_s}$ respectively, and the same goes for $\mu_{\hat{p_h}}$ and $\sigma_{\hat{p_h}}^2$. $\sigma_{\hat{p_s}\hat{p_h}}$ is the covariance of $\hat{p_s}$ and $\hat{p_h}$. $C_1$ and $C_2$ are two variables determined for stabilization. The former equals to $(k_1L)^2$, and the latter is $(k_2L)^2$. $L$ denotes the dynamic range of values in $I_{sr}$ and $I_{hr}$. $k_1$ and $k_2$ are two constants that are set as $0.01$ and $0.03$ respectively.

After all $m_s^i$ is computed, $M_s$ is produced through arranging all $m_s^i$ in a matrix according to their index. There are two benefits to measuring the similarity by similarity structure (SSIM). Firstly, SSIM is a perception-based criterion, and it is spatially stationary~\cite{wang2004image}. Secondly, SSIM is able to utilize its neighboring pixel values, which makes it more stable compared with absolute differences.

\subsection{Attention Map Calculation}
\label{subsec:amg}

\begin{figure}
	\centering
	\includegraphics[width=0.48\textwidth]{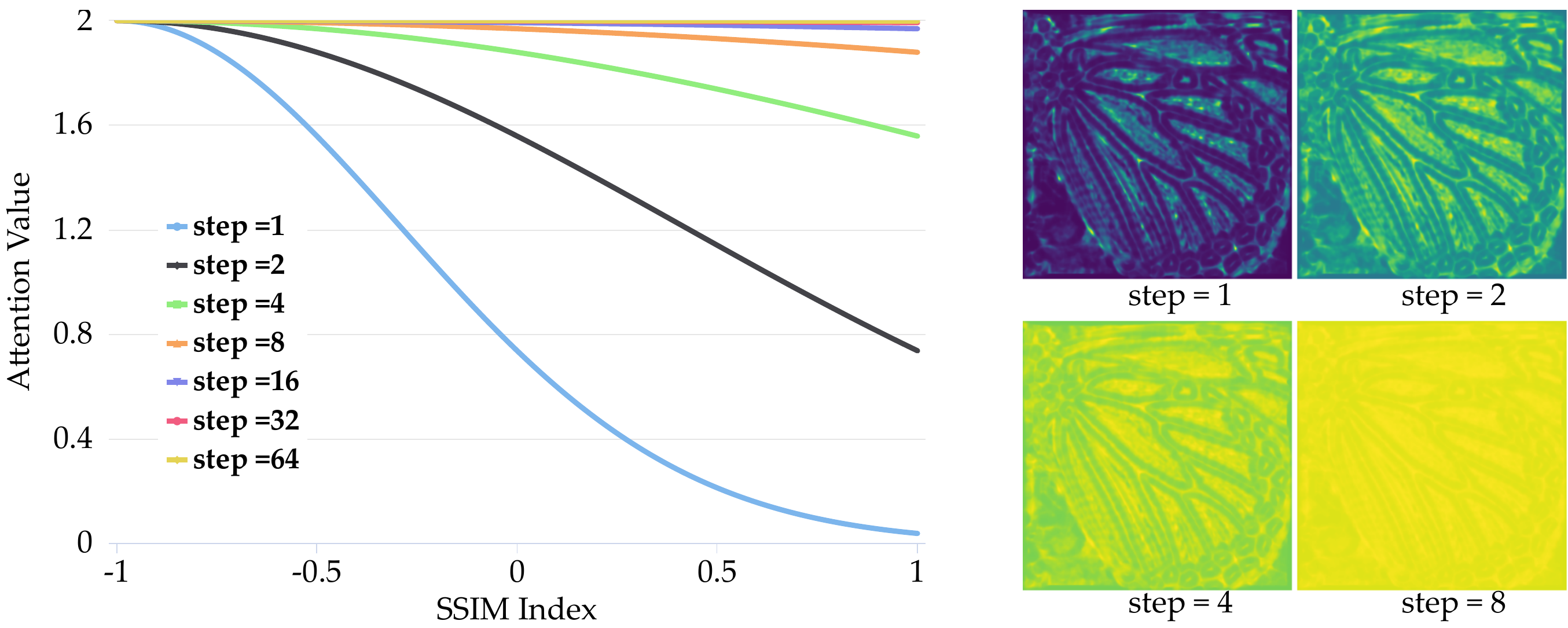}
    \caption{\textbf{An illustration of how $\mathbf{m_a}$ changes during training process.} The lines in the left chart display the distribution of $m_a$ in different step; these four blocks in the right exhibit an attention map example in different step.
    }
	\label{fig:gaussian}
\end{figure}

Attention map is the key point of PSPL. Given similarity map $M_s$ produced through the algorithm introduced in Sec.~\ref{subsec:psm}, its corresponding attention map $M_a$ is generated by a Gaussian function:
\begin{equation}
    M_a = \mathcal{G}_{\gamma, \mu, \delta}(M_s) \text{.} \label{eq:Ma}
\end{equation}
To be specific, for a value $m_s$ in $M_s$, its corresponding attention value $m_a$ is calculated by:
\begin{equation}
    m_a = \gamma \cdot \exp \left[-\frac {(m_s - \mu)^2} {\delta^2} \right] \text{.} \label{eq:ma}
\end{equation}
The distribution of Eq~(\ref{eq:ma}) illustrates a bell curve with $\gamma$ as its peak, $\mu$ as the position of the peak, and $\delta$ controlling its width. These three parameters determine various properties of $M_a$. Firstly, $\gamma$ is a nonnegative constant, which represents the maximal value that could obtained by Eq~(\ref{eq:ma}). Secondly, $\mu$ is a constant too, but it could be negative. According to Eq~(\ref{eq:ma}), if a similarity value $m_s$ is close to $\mu$, $m_a$ would be close to $\gamma$. In PSPL, $\mathcal{G}$ gives more attention to these pairs of pixels having smaller similarity. So $\mu$ equals the lower bound of the range of $m_s$, which is $-1$ because each $m_s$ is derived by SSIM Index. 
Different from the above two, $\delta$ is a variable that changes linearly depending on the training steps:
\begin{equation}
    \delta = \alpha \cdot \mathrm{step} + \beta \text{,} \label{eq:delta}
\end{equation}
in which $\alpha$ is growth rate, and $\beta$ is its initial value. Both of them are nonnegative constant. Obviously, $\delta$ is increased  gradually during training process, which prompts all values in $M_a$ to approach $\delta$. Fig.\ref{fig:gaussian} demonstrates how the relationship between $m_a$ and $\mathrm{step}$ changes in different training stage. In experiments, we set $\gamma = 2$, $\alpha = 1$ and $\beta = 0$.

Actually, the increase of $\delta$ leads to the decay of PSPL. In the beginning, the maximum and the minimum in $M_a$ have great difference, but all pixels would get the same attention after a certain growth. This degradation is inspired by self-paced learning~\cite{kumar2010self}, in which only easy samples are learned by learner in early, but all samples will be learned finally. However, PSPL is a pixel level self-paced learning. The attention value it generated determines which pixels should be learn first. 

\subsection{Loss Function}
\label{subsec:lf}

PSPL influences the optimization of models by affecting the loss function. For previous SISR models, their loss function can be formulated by:
\begin{equation}
    L(\Theta) = \mathcal{L}(I_{sr}, I_{hr}) \text{,} \label{eq:loss}
\end{equation}
in which $\Theta$ denotes the parameters in the trained model; $\mathcal{L}$ represents $\mathrm{L}1$ or $\mathrm{L}2$. The data flow of Eq~(\ref{eq:loss}) is shown by orange dash arrows in Fig.~\ref{fig:fosflow}.
While applying PSPL training strategy, the loss function is rewritten as:
\begin{equation}
    L(\Theta) = \mathcal{L}(M_a \odot I_{sr}, M_a \odot I_{hr}) \textbf{,} \label{eq:nloss}
\end{equation}
where $M_a$ is the attention map generated in Sec.~\ref{subsec:amg}, and $\odot$ represents element-wise multiplication. In Fig.~\ref{fig:fosflow}, how data flows in PSPL is displayed by green arrows. The green solid arrows represent the data flow when computing loss, and the green dash arrows represent how data flows when generating attention map.

Notably, the generation of $M_a$ is not a part of the network $\mathcal{F}_s$, which means both of $M_s$ and $M_a$ do not participate the backpropagation and can be seen as a constant matrix in each training step. In a sense, PSPL works like a teacher. Based on the local similarity between $I_{sr}$ and $I_{hr}$, it guides $\mathcal{L}$ to learn more from those pixels that differ widely. However, it only works during training, but does not appear in the test phase.

\section{Experiments}
\label{sec:exp}

The following contents are mainly divided into two parts. Firstly, some ablation experiments are conducted to exhibit the effect of PSPL. Secondly, the results of adopting PSPL on several existing outstanding models are compared with its original results. 

\begin{table*}[]
    \centering
    \caption{Comparison with the state-of-the-art}
    \label{tab:stot}
    \begin{tabular}{c!{\vrule width1.2pt}c!{\vrule width1.2pt}c|c!{\vrule width1.2pt}c|c!{\vrule width1.2pt}c|c!{\vrule width1.2pt}c|c!{\vrule width1.2pt}c|c}
    \Xhline{1.2pt}
    \multirow{2}{*}{Dataset}    & \multirow{2}{*}{Method}     & \multicolumn{2}{c!{\vrule width1.2pt}}{SRCNN~\cite{dong2014learning}} & \multicolumn{2}{c!{\vrule width1.2pt}}{VDSR~\cite{kim2016accurate}} & \multicolumn{2}{c!{\vrule width1.2pt}}{DRRN~\cite{tai2017image}} & \multicolumn{2}{c!{\vrule width1.2pt}}{LapSRN~\cite{LapSRN}} & \multicolumn{2}{c}{EDSR~\cite{lim2017enhanced}}             \\ \cline{3-12} 
                                &                             & PSNR                 & SSIM                   & PSNR            & SSIM                       & PSNR             & SSIM                      & PSNR              & SSIM                       & PSNR             & SSIM              \\ \Xhline{1.2pt}
    \multirow{2}{*}{Set5}       & baseline                    & 30.48                & 0.8628                 & 31.35           & 0.8838                     & 31.68            & 0.8888                    & 31.54             & 0.8850                     & 32.46            & 0.8968            \\ \cline{2-12} 
                                & PSPL                         & \textbf{30.64}       & \textbf{0.8692}        & \textbf{31.51}  & \textbf{0.8860}            & \textbf{31.73}   & \textbf{0.9041}           & \textbf{31.66}    & \textbf{0.9026}            & \textbf{32.51}   & \textbf{0.9008}   \\ \Xhline{1.2pt}
    \multirow{2}{*}{Set14}      & baseline                    & 27.49                & 0.7503                 & 28.02           & 0.7674                     & 28.21            & 0.7720                    & 28.19             & 0.7720                     & 28.80            & 0.7876            \\ \cline{2-12} 
                                & PSPL                         & \textbf{27.78}       & \textbf{0.7641}        & \textbf{28.29}  & \textbf{0.7779}            & \textbf{28.29}   & \textbf{0.7993}           & \textbf{28.29}    & \textbf{0.7984}            & \textbf{28.92}   & \textbf{0.7952}   \\ \Xhline{1.2pt}
    \multirow{2}{*}{B100}       & baseline                    & 26.90                & 0.7101                 & 27.29           & 0.7251                     & 27.38            & 0.7284                    & 27.32             & 0.7270                     & 27.71            & 0.7420            \\ \cline{2-12} 
                                & PSPL                         & \textbf{27.78}       & \textbf{0.7488}        & \textbf{28.12}  & \textbf{0.7593}            & \textbf{28.23}   & \textbf{0.7827}           & \textbf{28.23}    & \textbf{0.7820}            & \textbf{28.63}   & \textbf{0.7762}   \\ \Xhline{1.2pt}
    \multirow{2}{*}{Urban100}   & baseline                    & 24.52                & 0.7221                 & 25.18           & 0.7524                     & 25.44            & 0.7638                    & 25.21             & 0.7560                     & \textbf{26.64}   & 0.8033            \\ \cline{2-12} 
                                & PSPL                         & \textbf{24.57}       & \textbf{0.7292}        & \textbf{25.31}  & \textbf{0.7595}            & \textbf{25.54}   & \textbf{0.7854}           & \textbf{25.44}    & \textbf{0.7806}            & 26.63            & \textbf{0.8052}   \\ \Xhline{1.2pt}
    \end{tabular}
\end{table*}

\subsection{Ablation Experiments}
\label{subsec:ae}

This part introduces an ablation experiment we have conducted. In the experiment, a lightweight EDSR~\cite{lim2017enhanced} network is deployed to clarify how PSPL promotes the training efficiency and elevate the performance of SISR models. This simple EDSR only consists of 16 residual units, and each of them only outputs a feature map with 64 channels. Two models (with and without PSPL) are trained and validated with $\times 2$ scale on DIV2K~\cite{DBLP:conf/cvpr/AgustssonT17}, which consists of 800 training images and 100 validation images. 

\begin{figure}
	\centering
	\includegraphics[width=0.48\textwidth]{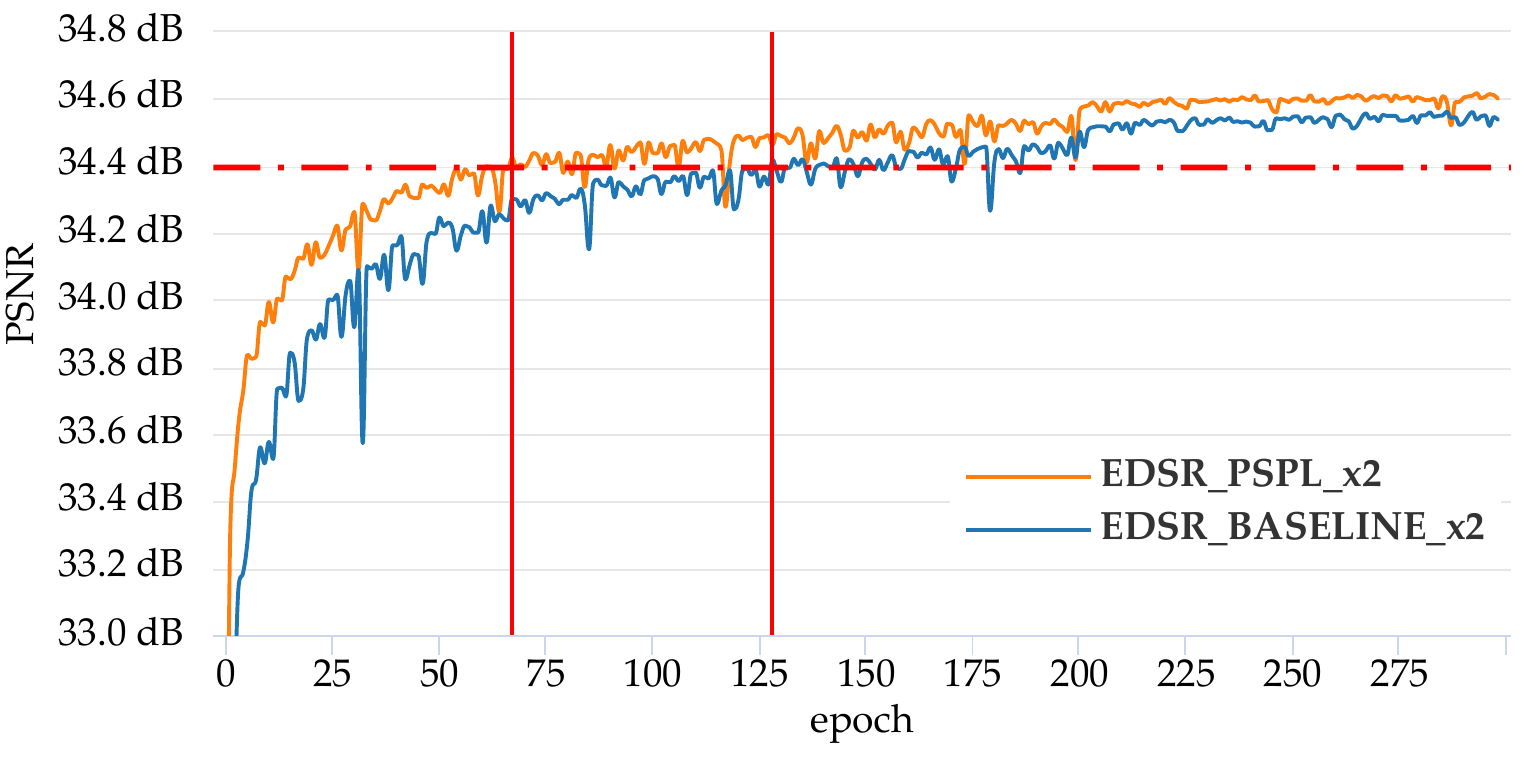}
    \caption{\textbf{Comparison of EDSR baseline model with and without PSPL.} }
	\label{fig:cebf}
\end{figure}

As shown in Fig.~\ref{fig:cebf}, both of them are trained and validated in 300 epochs.
However, EDSR+PSPL always performs better than the vanilla EDSR. Besides, the red plotlines also display how many epochs these two models need to train when they first achieve the PSNR of $34.4~\textrm{dB}$ respectively. EDSR+PSPL model reaches it in about 70 epochs, but the vanilla EDSR needs around 125 epochs. This simple experiment fully displays the superiority of PSPL.

\subsection{Comparison with Some Outstanding Methods}
\label{subsec:cwts}

Besides the ablation experiments, we also apply PSPL on several existing outstanding deep models in SISR. And This part discusses how they perform with and without PSPL. To be specific, we evaluate PSPL with 
SRCNN, VDSR, DRRN, LapSRN, and EDSR.
 All models in experiments are trained on DIV2K dataset and tested on Set5, Set14, B100, and Urban100. Detailed super parameters for a specific model still follow its original setting without any change.

\begin{figure}
	\centering
	\includegraphics[width=0.48\textwidth]{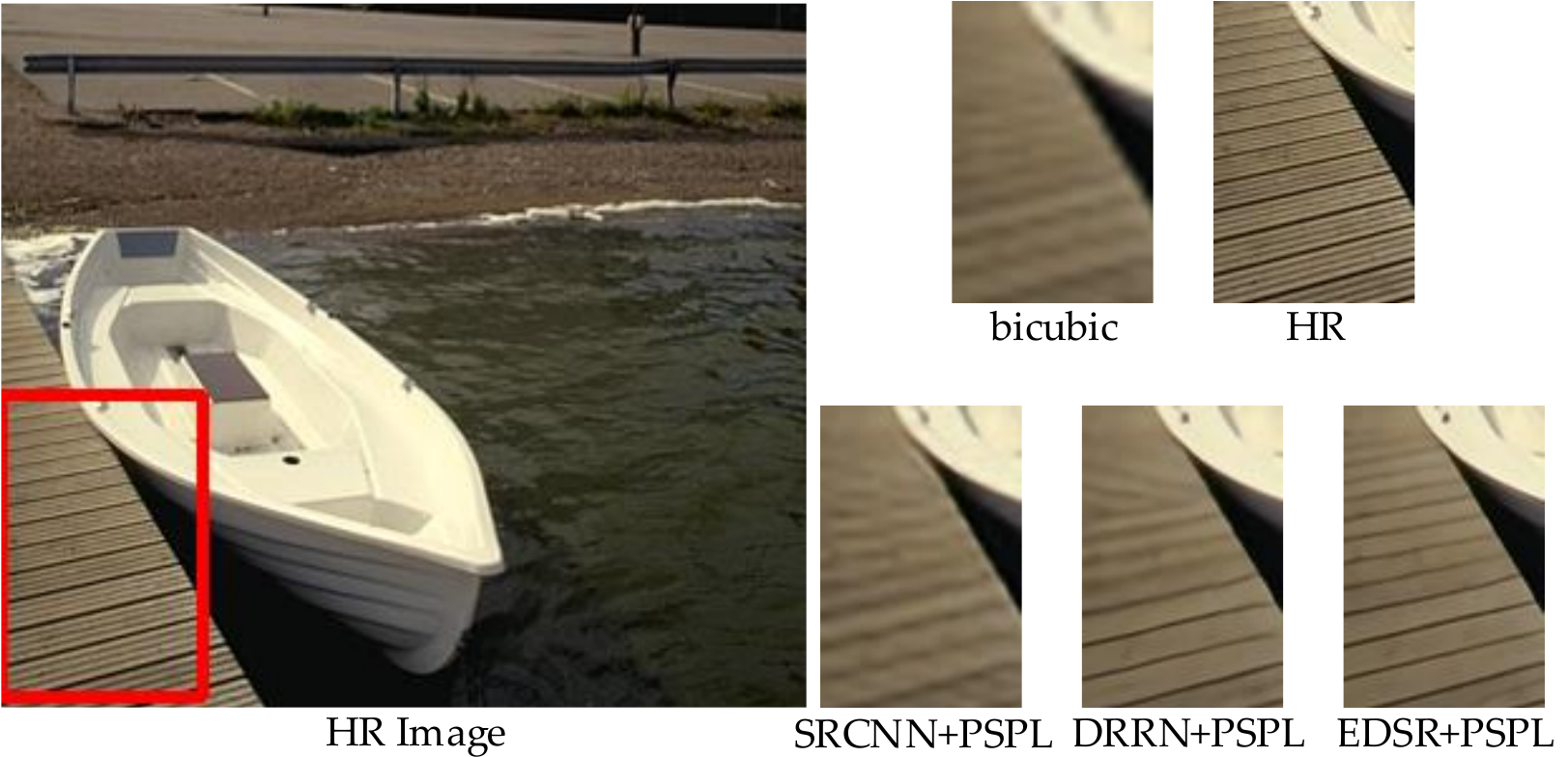}
    \caption{\textbf{A visual example.}}
	\label{fig:ave}
\end{figure}

On account of the limitation of this paper, We only display the results on $\times 4$ super-resolution. Fig.~\ref{fig:ave} displays some visual results, and Table.~\ref{tab:stot} lists the quantitative results of experimental models. From the table we can see, PSPL could improve their all performance to new heights. This illustrates that PSPL not only accelerates the training process, but also guide the parameters of trained models to a better parameter space and enhance their generalization. Even EDSR+PSPL does not perform better in Urban100 under PSNR, but it still produces better results under SSIM metric.


\section{Conclusion}
\label{sec:conc}

In this paper, we propose a training strategy to accelerate the training process and enhance the performance of SISR models called Pixel-level Self-Paced Learning (PSPL). It firstly generates an attention map according to the similarity between SR and HR image, and then grafts it onto loss function to impact the optimization of the trained SISR model. Moreover, all values in attention map is going to be close to the preset maximum value gradually during the training process, which is similar to the principle of self-paced learning. In the future, we would attempt to adopt PSPL on another pixel-wise regression task like depth estimation or deblurring, to further explore the ability of PSPL.

\bibliographystyle{IEEEbib}
\bibliography{refs}

\begin{thebibliography}{10}

\bibitem{wang2019transform}
Chunpeng Wang, Simiao Wang, Bin Ma, Jian Li, Xiangjun Dong, and Zhiqiu Xia,
\newblock ``Transform domain based medical image super-resolution via deep
  multi-scale network,''
\newblock in {\em ICASSP 2019-2019 IEEE International Conference on Acoustics,
  Speech and Signal Processing (ICASSP)}. IEEE, 2019, pp. 2387--2391.

\bibitem{huang2019hyperspectral}
Qian Huang, Wei Li, Ting Hu, and Ran Tao,
\newblock ``Hyperspectral image super-resolution using generative adversarial
  network and residual learning,''
\newblock in {\em ICASSP 2019-2019 IEEE International Conference on Acoustics,
  Speech and Signal Processing (ICASSP)}. IEEE, 2019, pp. 3012--3016.

\bibitem{yang2019deep}
Wenming Yang, Xuechen Zhang, Yapeng Tian, Wei Wang, Jing-Hao Xue, and Qingmin
  Liao,
\newblock ``Deep learning for single image super-resolution: A brief review,''
\newblock {\em IEEE Transactions on Multimedia}, 2019.

\bibitem{wang2014learning}
Qi~Wang and Yuan Yuan,
\newblock ``Learning to resize image,''
\newblock {\em Neurocomputing}, vol. 131, pp. 357--367, 2014.

\bibitem{wang2018detecting}
Qi~Wang, Mulin Chen, Feiping Nie, and Xuelong Li,
\newblock ``Detecting coherent groups in crowd scenes by multiview
  clustering,''
\newblock {\em IEEE transactions on pattern analysis and machine intelligence},
  vol. 42, no. 1, pp. 46--58, 2020.

\bibitem{DBLP:conf/icassp/ChangC19}
Chia{-}Yang Chang and Shao{-}Yi Chien,
\newblock ``Multi-scale dense network for single-image super-resolution,''
\newblock in {\em {IEEE} International Conference on Acoustics, Speech and
  Signal Processing, {ICASSP} 2019, Brighton, United Kingdom, May 12-17, 2019},
  2019, pp. 1742--1746.

\bibitem{DBLP:conf/icassp/BareLYFY18}
Bahetiyaer Bare, Ke~Li, Bo~Yan, Bailan Feng, and Chunfeng Yao,
\newblock ``A deep learning based no-reference image quality assessment model
  for single-image super-resolution,''
\newblock in {\em 2018 {IEEE} International Conference on Acoustics, Speech and
  Signal Processing, {ICASSP} 2018, Calgary, AB, Canada, April 15-20, 2018},
  2018, pp. 1223--1227.

\bibitem{dong2014learning}
Chao Dong, Chen~Change Loy, Kaiming He, and Xiaoou Tang,
\newblock ``Learning a deep convolutional network for image super-resolution,''
\newblock in {\em Computer Vision -- ECCV 2014}, Cham, 2014, pp. 184--199,
  Springer International Publishing.

\bibitem{LapSRN}
Wei-Sheng Lai, Jia-Bin Huang, Narendra Ahuja, and Ming-Hsuan Yang,
\newblock ``Deep laplacian pyramid networks for fast and accurate
  super-resolution,''
\newblock in {\em IEEE Conference on Computer Vision and Pattern Recognition},
  2017.

\bibitem{kim2016accurate}
Jiwon Kim, Jung Kwon~Lee, and Kyoung Mu~Lee,
\newblock ``Accurate image super-resolution using very deep convolutional
  networks,''
\newblock in {\em Proceedings of the IEEE conference on computer vision and
  pattern recognition}, 2016, pp. 1646--1654.

\bibitem{tai2017memnet}
Ying Tai, Jian Yang, Xiaoming Liu, and Chunyan Xu,
\newblock ``Memnet: A persistent memory network for image restoration,''
\newblock in {\em Proceedings of the IEEE international conference on computer
  vision}, 2017, pp. 4539--4547.

\bibitem{lim2017enhanced}
Bee Lim, Sanghyun Son, Heewon Kim, Seungjun Nah, and Kyoung Mu~Lee,
\newblock ``Enhanced deep residual networks for single image
  super-resolution,''
\newblock in {\em Proceedings of the IEEE conference on computer vision and
  pattern recognition workshops}, 2017, pp. 136--144.

\bibitem{haris2018deep}
Muhammad Haris, Gregory Shakhnarovich, and Norimichi Ukita,
\newblock ``Deep back-projection networks for super-resolution,''
\newblock in {\em Proceedings of the IEEE conference on computer vision and
  pattern recognition}, 2018, pp. 1664--1673.

\bibitem{tai2017image}
Ying Tai, Jian Yang, and Xiaoming Liu,
\newblock ``Image super-resolution via deep recursive residual network,''
\newblock in {\em Proceedings of the IEEE conference on computer vision and
  pattern recognition}, 2017, pp. 3147--3155.

\bibitem{tang2016example}
Yi~Tang, Hong Chen, Zhanwen Liu, Biqin Song, and Qi~Wang,
\newblock ``Example-based super-resolution via social images,''
\newblock {\em Neurocomputing}, vol. 172, pp. 38--47, 2016.

\bibitem{ledig2017photo}
Christian Ledig, Lucas Theis, Ferenc Husz{\'a}r, Jose Caballero, Andrew
  Cunningham, Alejandro Acosta, Andrew Aitken, Alykhan Tejani, Johannes Totz,
  Zehan Wang, et~al.,
\newblock ``Photo-realistic single image super-resolution using a generative
  adversarial network,''
\newblock in {\em Proceedings of the IEEE conference on computer vision and
  pattern recognition}, 2017, pp. 4681--4690.

\bibitem{wang2004image}
Zhou Wang, Alan~C Bovik, Hamid~R Sheikh, Eero~P Simoncelli, et~al.,
\newblock ``Image quality assessment: from error visibility to structural
  similarity,''
\newblock {\em IEEE transactions on image processing}, vol. 13, no. 4, pp.
  600--612, 2004.

\bibitem{kumar2010self}
M~Pawan Kumar, Benjamin Packer, and Daphne Koller,
\newblock ``Self-paced learning for latent variable models,''
\newblock in {\em Advances in Neural Information Processing Systems}, 2010, pp.
  1189--1197.

\bibitem{DBLP:conf/cvpr/AgustssonT17}
Eirikur Agustsson and Radu Timofte,
\newblock ``{NTIRE} 2017 challenge on single image super-resolution: Dataset
  and study,''
\newblock in {\em 2017 {IEEE} Conference on Computer Vision and Pattern
  Recognition Workshops, {CVPR} Workshops 2017, Honolulu, HI, USA, July 21-26,
  2017}, 2017, pp. 1122--1131.

\end{thebibliography}

\end{document}